\documentclass[10pt,twocolumn,letterpaper]{article}

\usepackage{cvpr}
\usepackage{times}
\usepackage{epsfig}
\usepackage{graphicx}
\usepackage{amsmath}
\usepackage{amssymb}

\usepackage{enumitem}
\usepackage{subfig}
\usepackage{floatrow}
\usepackage{multirow}

\usepackage{fancyhdr}

\renewcommand\etal[1]{\textit{et al.}~\cite{#1}}

\newcommand\ic[2][1]{\frame{\includegraphics[width=#1\textwidth]{#2}}}
\renewcommand\cap[3]{\caption[#2]{\label{#1}\textsc{#2} \small{#3}}}

\usepackage[pagebackref=true,breaklinks=true,letterpaper=true,colorlinks,bookmarks=false]{hyperref}

\cvprfinalcopy % *** Uncomment this line for the final submission

\def\cvprPaperID{0010} % *** Enter the CVPR Paper ID here

\ifcvprfinal\fi
\begin{document}

%%%%%%%%% TITLE
\title{Adversarial Attack on Deep Learning-Based Splice Localization}

\author{Andras Rozsa\\
Verisk Analytics\\
Jersey City, NJ\\
{\tt\small andras.rozsa@verisk.com}
\and
Zheng Zhong\\
Verisk Analytics\\
Jersey City, NJ\\
{\tt\small zheng.zhong@verisk.com}
\and
Terrance E. Boult\\
UCCS VAST Lab\\
Colorado Springs, CO\\
{\tt\small tboult@vast.uccs.edu}
}

\maketitle
\thispagestyle{empty}

% ***** For arXiv *****
\chead{\footnotesize This is a pre-print of the original paper accepted at the CVPR Workshop on Media Forensics 2020}
\thispagestyle{fancy}
\pagenumbering{gobble}
% *********************

%%%%%%%%% ABSTRACT
\begin{abstract}
Regarding image forensics, researchers have proposed various approaches to detect and/or localize manipulations, such as splices.
Recent best performing image-forensics algorithms greatly benefit from the application of deep learning, but such tools can be vulnerable to adversarial attacks.
Due to the fact that most of the proposed adversarial example generation techniques can be used only on end-to-end classifiers, the adversarial robustness of image-forensics methods that utilize deep learning only for feature extraction has not been studied yet. 
Using a novel algorithm capable of directly adjusting the underlying representations of patches we demonstrate on three non end-to-end deep learning-based splice localization tools that hiding manipulations of images is feasible via adversarial attacks.
While the tested image-forensics methods, EXIF-SC, SpliceRadar, and Noiseprint, rely on feature extractors that were trained on different surrogate tasks, we find that the formed adversarial perturbations can be transferable among them regarding the deterioration of their localization performance.
\end{abstract}

%%%%%%%%% BODY TEXT

\section{Introduction}

There is an increasing demand for tools capable of detecting manipulated digital media, including fake pictures.
Researchers proposed various approaches to detect and/or localize image manipulations.
While detection only provides a verdict on the authenticity of the picture in question, localization shows the tampered pixels of the manipulated image and, therefore, gives us a better insight about the nature and goal of the modification.
Recent advances in image forensics are, at least, partially due to the application of machine learning as some of the best performing splice localization systems are deep learning-based approaches \cite{bappy2017exploiting,huh2018fighting,ghosh2019spliceradar,cozzolino2019noiseprint,zhou2018learning,salloum2018image,Wu_2019_CVPR}.

\begin{figure}
    \centering
    \includegraphics[width=0.99\columnwidth]{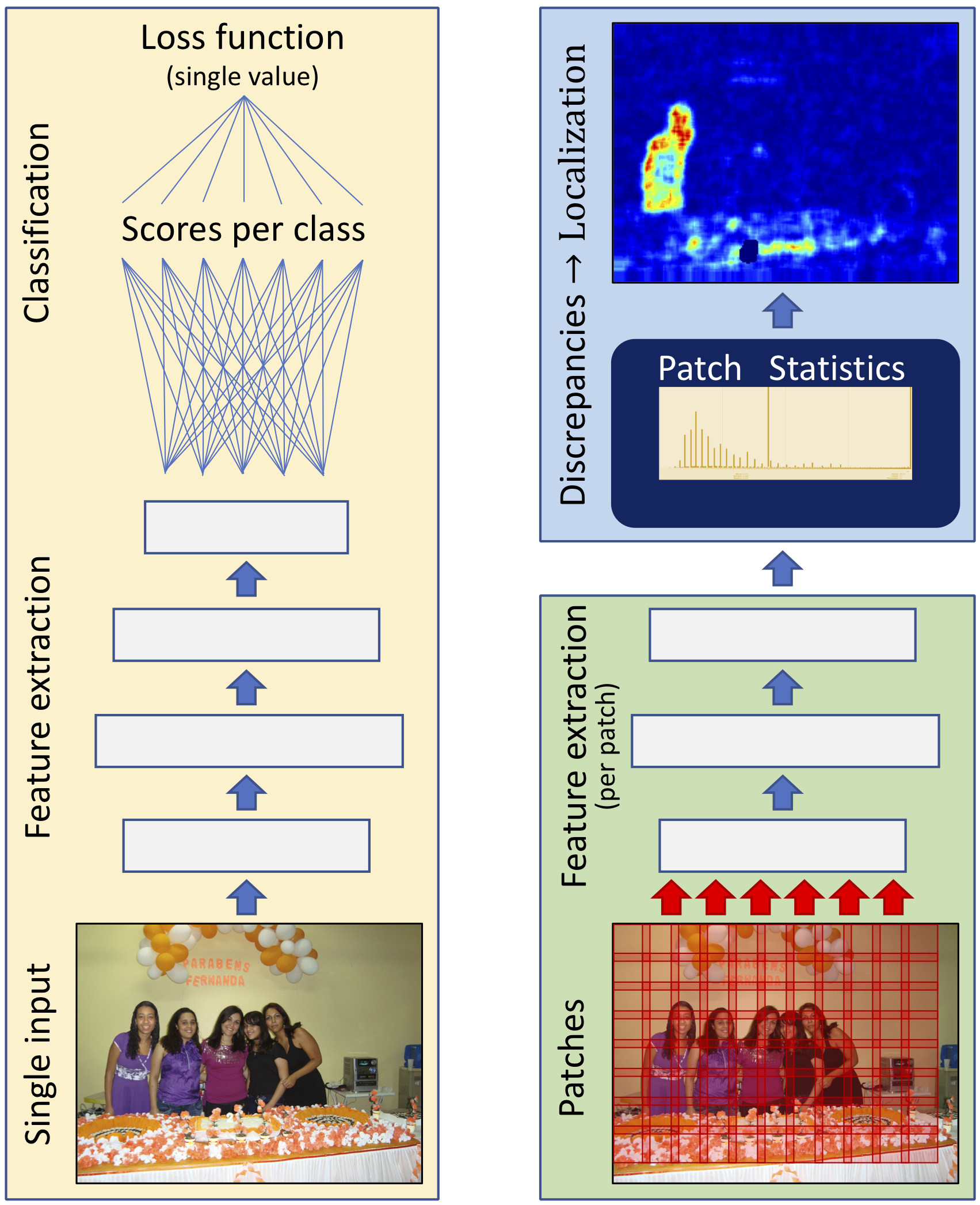}
    \cap{fig:teaser}{Adversarial Example generation.}{Adversarial attacks on networks performing recognition tasks, as shown on the left, are simple as there is one loss function that combines the scores from the classes and it can be back-propagated to the input image. Splice localization tools, displayed on the right, break their input image into 100s or 1000s of patches that often overlap. Then a feature representation is extracted for each patch, from which group statistics identify discrepancies yielding the localization of manipulated regions.  In such systems there is no single loss function to target and there is also no differential path backwards through the statistical processes. This paper develops an alternative attack by directly manipulating the patch representations.}
    \vspace{-3pt}
\end{figure}

Although the utilization of deep learning yields improvements in image forensics, deep learning models can be manipulated.
Szegedy \etal{szegedy2013intriguing} demonstrated that machine learning models, including the state-of-the-art deep neural networks, can misclassify slightly perturbed, otherwise correctly classified inputs.
The existence of such adversarial examples raises fundamental questions about the generalization properties and the real-world applications of machine learning models.

Researchers developed various counter-forensics tools to impede traditional multimedia-forensics methods -- Barni \etal{barni2018adversarial} provides a detailed overview on the subject area -- but the adversarial robustness of the latest deep learning-based manipulation localization approaches has not been analyzed yet.
While the latest image-forensics methods utilize deep learning models for extracting feature representations to localize manipulations, most adversarial attacks are capable of forming adversarial perturbations only on end-to-end recognition systems \cite{szegedy2013intriguing, goodfellow2014explaining,kurakin2017adversarial,carlini2017towards,papernot2016limitations,madry2018towards,dabouei2019fast}.
Consequently, as visually demonstrated in Figure \ref{fig:teaser}, such approaches are not suitable for performing adversarial attacks on non end-to-end systems.

In this paper, we show that non end-to-end deep learning-based splice localization tools can be manipulated via adversarial attacks.
We perform experiments on three of the best performing tools that are available, EXIF-SC \cite{huh2018fighting}, SpliceRadar \cite{ghosh2019spliceradar}, and Noiseprint \cite{cozzolino2019noiseprint}.
Our experiments on three datasets demonstrate that the splice localization performance of the targeted deep learning-based image-forensics tools can significantly deteriorate due to our novel adversarial attack.
Namely, by imperceptibly and simultaneously perturbing each image patch, we can hide manipulations of spliced images from the targeted image-forensics methods.
We find that adversarial perturbations between the three image-forensics tools can be transferable, with different effectiveness.
For instance, the adversarial examples generated on EXIF-SC tend to greatly decrease the localization performance of SpliceRadar.
Interestingly, this transferability appears to be asymmetric as it does not occur in the other direction.
Overall, our experimental results suggest that Noiseprint is the most robust among the three tested image-forensics tools.
\section{Related Work}

Many image-forensics techniques, especially manipulation detection and localization approaches, rely on artifacts that are introduced by the image formation pipeline. 
The idea is that manipulations, although they may not leave any visual clues, alter such underlying artifacts of images \cite{popescu2004statistical}.

EXIF-SC proposed by Huh \etal{huh2018fighting} utilizes EXIF metadata for manipulation detection and localization.
Given a pair of image patches, a Siamese network has been trained to predict the probability that patches share the same value for each of the 83 EXIF metadata attributes. 
Each branch of the network produces a 4096-dimensional feature vector which represents the EXIF metadata information of the particular patch.
Following that, a two-layer Multi-Layer Perceptron (MLP) is applied to combine the EXIF metadata consistency predictions to an overall consistency. 
Finally, mean shift \cite{cheng1995mean} is used for finding the most consistent mode among all consistency maps of the selected patches.

SpliceRadar of Ghosh \etal{ghosh2019spliceradar} relies on low-level image statistics related to camera models and disregards high-level features connected to semantics.
The feature extractor has been trained via a surrogate task of camera model identification.
After an image patch is fed into a learned and constrained convolutional layer mimicking the filters proposed in \cite{fridrich2012rich}, mutual information computed between the patch and its corresponding intermediate feature representation is used as a penalty to further suppress semantics.
During inference, a 100-dimensional feature representation is obtained for each image patch. 
Similar to \cite{cozzolino2015splicebuster}, a Gaussian Mixture Model (GMM) with two components is applied to localize genuine and forged regions of the particular image based upon the feature representation of the patches. 

Noiseprint proposed by Cozzolino \etal{cozzolino2019noiseprint} extracts noise residuals -- called noiseprints -- from images and utilizes those patterns as camera model fingerprints.
Based on a convolutional image denoiser proposed in \cite{zhang2017beyond}, the authors train a Siamese network to classify whether a pair of patches originate from the same camera model and location. 
Each branch of the network yields a 2-D feature map which represents the camera noise residuals.
A distance-based logistic loss is applied during training to ensure small distances between noiseprints that belong to same camera model/location and large distances otherwise. 
During inference, noiseprint is extracted via the trained image denoiser for the particular image. 
With the assumption that genuine and forged patches originate from different camera models and, therefore, yield discrepancies on noise residuals of the corresponding patches, Noiseprint, similar to \cite{cozzolino2015splicebuster}, utilizes a GMM to separate genuine and manipulated regions.

The feasibility and effectiveness of adversarial attacks as counter-forensics on non end-to-end deep learning-based manipulation localization approaches, such as EXIF-SC, SpliceRadar, or Noiseprint, have not been analyzed yet.
G{\"u}era \etal{guera2017counter} showed that an end-to-end Convolutional Neural Network (CNN), which was trained to identify the camera source of images, can be spoofed via adversarial attacks, such as the Fast Gradient Sign (FGS) method \cite{goodfellow2014explaining} or the Jacobian-based Saliency Map Attack (JSMA) \cite{papernot2016limitations}.
Barni \etal{barni2019transferability} assessed the transferability of adversarial examples formed via the iterative FGS \cite{kurakin2017adversarial} and JSMA methods using two CNN architectures that were trained to detect image resizing and median filtering, respectively.
Note that the two previously mentioned related works experiment with adversarial attacks on end-to-end networks.
Also, those image-forensics algorithms perform manipulation detection rather than localization.

To perform adversarial attacks on EXIF-SC, SpliceRadar, or Noiseprint, adversaries need to directly manipulate the internal feature representations these image-forensics methods rely on.
The approach of Sabour \etal{sabour2016adversarial} yields adversarial examples that cause misclassifications and also mimic the internal representations of targeted inputs.
They use the computationally expensive L-BFGS optimization technique, which limits the application of their technique.
The Layerwise Origin-Target Synthesis (LOTS) of Rozsa \etal{rozsa2017bmvc,rozsa2017lots} is similar as it is also capable of directly manipulating internal representations, but it is computationally more efficient.
Inkawhich \etal{inkawhich2019feature} used an approach very similar to LOTS to show its effectiveness of forming transferable adversarial examples between end-to-end deep learning-based recognition systems.
To experiment with the targeted image-forensics methods, we adapt LOTS to simultaneously manipulate the feature representations of overlapping patches representing the spliced images.

\section{Approach}
\label{approach}

While the targeted image-forensics algorithms solve the same localization task on manipulated images very differently, from a bird's-eye view, their architectures show some similarities.
Namely, they extract feature representations from patches and utilize those for splice localization.
While SpliceRadar and Noiseprint simply perform clustering based upon the extracted feature representations using a two-component Gaussian Mixture Model (GMM), EXIF-SC derives consistency maps for each patch using their feature representations and combines those to identify discrepancies.
Consequently, if an adversary is able to modify pixels of patches and make feature representations of those similar, these image-forensics tools will fail to differentiate authentic and manipulated regions.

To be able to perform adversarial attacks on the targeted image-forensics methods, we adapt LOTS \cite{rozsa2017bmvc,rozsa2017lots} to simultaneously manipulate the feature representations of image patches that these splice localization tools rely on.
Considering a manipulated image $X$ and its corresponding set of patches $P$, we start with identifying authentic patches, denoted as $A$, from $P$.
We define a patch authentic if and only if each of its pixels are authentic.
Next, we specify the target representation $t$ to be the mean feature representation of authentic patches.
Our goal is to make both authentic and manipulated patches mimic $t$ with their feature representations.
To do so, for each patch we utilize a Euclidean loss defined on its feature representation $f\left(P_i\right)$ and the target representation $t$, and apply its gradient with respect to the pixels of patch $P_i$ to move its feature representation closer to $t$. These patch-gradients can be defined as
\begin{equation}
\label{lots_grad}
  \eta_i(P_i,t) = {{\nabla}_{P_i}} \left( \frac{1}{2} \left\| t - f(P_i) \right\|^2\right), \quad i=0,...,L \,\,,
\end{equation}
where $L$ is the number of patches for image $X$.
Note that patches ($P$) and feature representations of patches ($f \left(P_i\right)$) are obtained as defined by the targeted image-forensics tool.

The targeted image-forensics methods can rely on multiple, sometimes overlapping, patches to localize manipulated regions of images.
Therefore, we simply combine the calculated patch-gradients $\eta_i(P_i,t)$ according to their corresponding locations on the particular image $X$ to obtain $G\left(P,t\right)$.
Note that $G$ has the same dimensions as $X$.
We calculate and apply the combined gradients iteratively.
Therefore, our algorithm can be formalized as 
\begin{equation}
\label{lots_iter}
  X_0 = X, \quad X_{n+1} = X_n - \alpha \frac{G\left(P^{(n)},t\right)}{\left\|G\left(P^{(n)},t\right)\right\|_\infty},
\end{equation}
where $P^{(n)}$ refers to the patches of $X_n$, and $\alpha$ defines the step-size.
After applying the scaled and combined gradients to $X_n$, we clip pixels to be in the range of $\left[0,255\right]$.
Note that the number of patches and their locations remain the same, while pixel-values for $P^{(n)}$ can vary between iterations.

In our experiments, we run this algorithm for a predefined number of iterations with a fixed step-size $\alpha$, and choose the perturbed image $X_n$ with patches having the smallest average distance from target $t$.
After identifying the best, we round pixel-values and save the resulting perturbed image.

\begin{table*}[ht]
\footnotesize
\setlength{\tabcolsep}{6pt}
\begin{center}
\begin{tabular}{|p{1pt}l|cccc|cccc|cccc|}
\hline
\multicolumn{2}{|l|}{LOCALIZATION METHOD} & \multicolumn{4}{c|}{COLUMBIA} & \multicolumn{4}{c|}{DSO-1} & \multicolumn{4}{c|}{NC16}\\
\multicolumn{2}{|l|}{(results on spliced images)} & F1 & MCC & AUC & MAP & F1 & MCC & AUC & MAP & F1 & MCC & AUC & MAP\\
\hline\hline
\multirow{3}{*}{\rotatebox[origin=c]{90}{\footnotesize{BL}}} &
EXIF-SC & 0.911 & 0.878 & 0.980 & 0.952 & 0.568 & 0.517 & 0.845 & 0.526 & 0.377 & 0.359 & 0.796 & 0.340 \\
& SpliceRadar & 0.785 & 0.698 & 0.877 & 0.758 & 0.690 & 0.656 & 0.912 & 0.657 & 0.399 & 0.382 & 0.813 & 0.349 \\
& Noiseprint & 0.816 & 0.736 & 0.905 & 0.817 & 0.800 & 0.782 & 0.928 & 0.775 & 0.415 & 0.404 & 0.781 & 0.352 \\ 
\hline
\multirow{3}{*}{\rotatebox[origin=c]{90}{\footnotesize{WBA}}} &
EXIF-SC & 0.696 & 0.553 & 0.812 & 0.688 & 0.350 & 0.249 & 0.691 & 0.266 & 0.327 & 0.308 & 0.769 & 0.281 \\
& SpliceRadar & 0.583 & 0.365 & 0.691 & 0.498 & 0.362 & 0.274 & 0.687 & 0.273 & 0.305 & 0.289 & 0.756 & 0.246 \\
& Noiseprint  & 0.748 & 0.636 & 0.859 & 0.769 & 0.573 & 0.515 & 0.812 & 0.506 & 0.332 & 0.311 & 0.730 & 0.273 \\ 
\hline
\multirow{6}{*}{\rotatebox[origin=c]{90}{\footnotesize{TRA}}} &
EXIF-SC on SpliceRadar Ex. & 0.896 & 0.860 & 0.973 & 0.937 & 0.557 & 0.507 & 0.845 & 0.516 & 0.378 & 0.360 & 0.795 & 0.341 \\
& EXIF-SC on Noiseprint Ex. & 0.887 & 0.849 & 0.970 & 0.933 & 0.556 & 0.502 & 0.840 & 0.513 & 0.375 & 0.360 & 0.797 & 0.338 \\ 
& SpliceRadar on EXIF-SC Ex. & 0.583 & 0.364 & 0.692 & 0.498 & 0.569 & 0.524 & 0.859 & 0.515 & 0.344 & 0.330 & 0.786 & 0.291 \\
& SpliceRadar on Noiseprint Ex. & 0.719 & 0.587 & 0.825 & 0.681 & 0.676 & 0.640 & 0.903 & 0.637 & 0.389 & 0.367 & 0.807 & 0.332 \\ 
& Noiseprint on EXIF-SC Ex. & 0.729 & 0.610 & 0.855 & 0.746 & 0.679 & 0.638 & 0.876 & 0.648 & 0.321 & 0.301 & 0.747 & 0.264 \\ 
& Noiseprint on SpliceRadar Ex. & 0.702 & 0.564 & 0.818 & 0.670 & 0.735 & 0.704 & 0.911 & 0.720 & 0.355 & 0.335 & 0.763 & 0.304 \\ 
\hline
& \textit{All Authentic Attack} & 0.570 & 0.276 & N/A & N/A & 0.279 & 0.136 & N/A & N/A & 0.174 & 0.125 & N/A & N/A \\
\hline
\end{tabular}
\vspace{-8pt}
\end{center}

\cap{tab:results}{Quantitative Results.}{This table shows performance measures for EXIF-SC, SpliceRadar, and Noiseprint on the manipulated images of the Columbia, DSO-1, and NC16 datasets, respectively.
We show the baseline performances (BL) and how the manipulation localization performances of the targeted image-forensics methods change due to adversarial white-box attacks (WBA) and the transferability (TRA) of the formed adversarial perturbations.
The last row (All Authentic Attack) presents the applicable measures for ideal attacks where each pixel is declared authentic by the targeted image-forensics tool.}

    %\vspace{8pt}
    \vspace{-4pt}

\end{table*}

\section{Experiments}

In this section, after describing the experimental setup, we present the results of our adversarial attacks on three non end-to-end deep learning-based image-forensics methods.

\subsection{Experimental setup}

To help reproducability and future comparisons, we aim at revealing all details about our experiments and justifying various decisions we made for the experimental setup.

\textbf{Targeted image-forensics algorithms.}
Due to the nature of white-box attacks, our selection is limited to approaches with available source code.
Consequently, we choose EXIF-SC \cite{huh2018fighting}, SpliceRadar \cite{ghosh2019spliceradar}, and Noiseprint \cite{cozzolino2019noiseprint}.
As the Columbia dataset contains smaller images than DSO-1, we slightly adjust SpliceRadar for the former dataset by changing the stride of patches from $48$ to $24$ pixels.
This modification allows SpliceRadar to perform better due to the increased number of patches via the smaller stride.
Regarding the adapted LOTS, while attacking EXIF-SC and SpliceRadar, we aim at mimicking the mean feature representations of authentic patches with the 4096- and 100-dimensional feature vectors of patches, respectively.
The patch sizes and their locations are identical to those used for inference.
For Noiseprint -- based on preliminary experiments -- we utilize $8\times8$ pixel, non-overlapping patches for both defining the target as the mean feature representation of authentic patches and then moving feature representations of all patches closer to the specified target.
Other than the aforementioned modifications, we use the three image-forensics methods with their default settings.

\textbf{Datasets.}
Regarding the publicly available datasets that are commonly used in the literature for evaluating image-forensics tools, we choose two on which the targeted image-forensics methods perform well.
Therefore, forming adversarial perturbations for these spliced images is more challenging.
These are the Columbia \cite{hsu06crfcheck} and DSO-1 \cite{de2013exposing} datasets containing 180 and 100 spliced images, respectively, along with manipulation masks.
Comparing the two datasets, we would like to highlight that the images of Columbia contain larger, very perceptible spliced regions.
Note that we have excluded $5$ of the $100$ DSO-1 images from our experiments as the dimensions of the spliced images and their corresponding masks differ.
In addition to Columbia and DSO-1, we also run experiments on the more challenging NC16\footnote{\url{https://www.nist.gov/itl/iad/mig/nimble-challenge-2017-evaluation}} dataset which was collected for the Media Forensics Challenge\footnote{\url{https://www.nist.gov/itl/iad/mig/media-forensics-challenge}}.
Compared to Columbia or DSO-1, NC16 is a relatively large dataset containing 564 images with a great variety regarding image sizes, sizes of manipulated regions, acquired camera models, etc. 

\begin{table}[t]
\footnotesize
\setlength{\tabcolsep}{5pt}
\begin{center}
\begin{tabular}{|p{1pt}l|cccc|}
\hline
\multicolumn{2}{|l|}{LOCALIZATION METHOD} & \multicolumn{4}{c|}{NC16 SUBSET}\\
\multicolumn{2}{|l|}{(results on spliced images)} & F1 & MCC & AUC & MAP\\
\hline\hline
\multirow{3}{*}{\rotatebox[origin=c]{90}{\footnotesize{BL}}} &
EXIF-SC & 0.843 & 0.746 & 0.925 & 0.864 \\
& SpliceRadar & 0.823 & 0.734 & 0.912 & 0.827 \\
& Noiseprint & 0.837 & 0.754 & 0.868 & 0.799 \\ 
\hline
\multirow{3}{*}{\rotatebox[origin=c]{90}{\footnotesize{WBA}}} &
EXIF-SC & 0.644 & 0.542 & 0.851 & 0.631 \\
& SpliceRadar & 0.602 & 0.474 & 0.809 & 0.569 \\
& Noiseprint & 0.721 & 0.635 & 0.822 & 0.669 \\ 
\hline
\multirow{6}{*}{\rotatebox[origin=c]{90}{\footnotesize{TRA}}} &
EXIF-SC on SpliceRadar Ex. & 0.739 & 0.670 & 0.904 & 0.733 \\
& EXIF-SC on Noiseprint Ex. & 0.676 & 0.603 & 0.845 & 0.646 \\ 
& SpliceRadar on EXIF-SC Ex. & 0.803 & 0.719 & 0.917 & 0.821 \\
& SpliceRadar on Noiseprint Ex. & 0.729 & 0.646 & 0.859 & 0.705 \\ 
& Noiseprint on EXIF-SC Ex. & 0.793 & 0.714 & 0.917 & 0.810 \\ 
& Noiseprint on SpliceRadar Ex. & 0.764 & 0.678 & 0.898 & 0.760 \\ 
\hline
& \textit{All Authentic Attack} & 0.387 & 0.234 & N/A & N/A \\
\hline
\end{tabular}
\vspace{-3pt}
\end{center}

\cap{tab:nc16_sub}{Quantitative Results on NC16-Subset.}{This table shows performance measures on the subset of the NC16 dataset consisting of manipulated images for which EXIF-SC, SpliceRadar, and Noiseprint yield F1 scores greater than or equal to 0.6 (83 images).
We show the baseline performances (BL), performances of the targeted image-forensics tools under adversarial white-box attacks (WBA), and the transferability (TRA) of the formed adversarial perturbations.
The last row (All Authentic Attack) presents the applicable measures for ideal attacks where each pixel is declared authentic by the targeted image-forensics method.}

    %\vspace{8pt}
    \vspace{-6pt}

\end{table}

\textbf{Attacks.}
We utilize the algorithm that we described in Section \ref{approach}.
Based on small-scale experiments, we have chosen step-size of $5$ pixels ($\alpha=5$) and iterations to be limited to $50$.
Limiting the number of iterations naturally requires a larger step-size to be able to mimic the predefined target.
Furthermore, with a larger step-size, it is less likely that the algorithm will get stuck in a local optima regarding the surface of the Euclidean loss.
While these settings might yield sub-optimal results, we can limit the computational costs and evaluate the effectiveness of a general attack.
Note that due to our definition of an authentic patch and the default patch size of EXIF-SC (128$\times$128 pixels), four images of the NC16 dataset do not have a single authentic patch and, therefore, cannot be attacked with our algorithm.

\begin{figure*}
    \centering
    
    \subfloat[][\label{col_imgs:a}\centering Left to right: spliced image, ground-truth manipulation mask, heatmaps via EXIF-SC, SpliceRadar, and Noiseprint]{
    %\hspace{-5pt}
    \ic[.19]{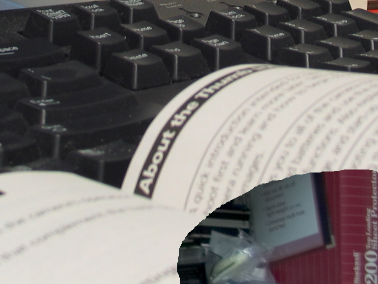} \hspace{-1pt} \ic[.19]{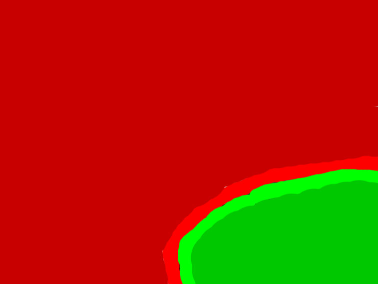} \hspace{-1pt} \ic[.19]{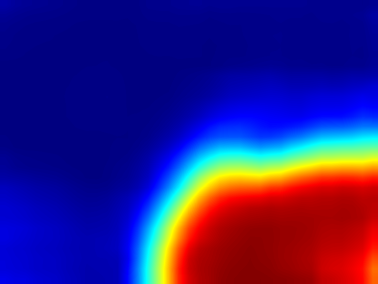} \hspace{-1pt}
    \ic[.19]{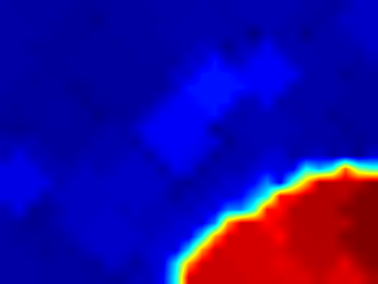} \hspace{-1pt}
    \ic[.19]{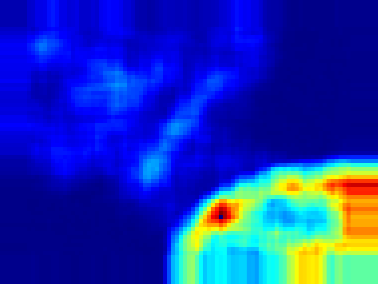}
    } \\
    \vspace{-5pt}
    
    \subfloat[][\label{col_imgs:b}\centering Left to right: adversarial image formed on EXIF-SC, visualized adversarial perturbation, heatmaps via EXIF-SC, SpliceRadar, and Noiseprint]{
    %\hspace{-5pt}
    \ic[.19]{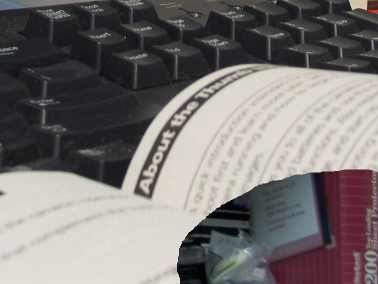} \hspace{-1pt}
    \ic[.19]{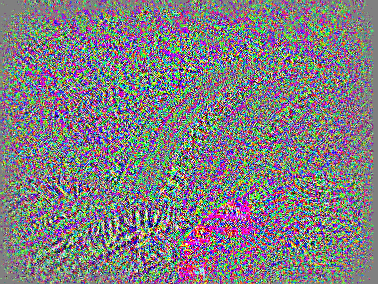} \hspace{-1pt}
    \ic[.19]{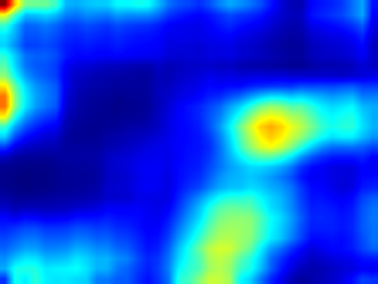} \hspace{-1pt}
    \ic[.19]{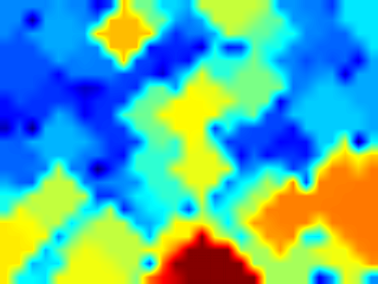} \hspace{-1pt}
    \ic[.19]{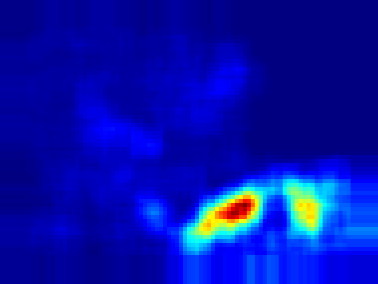}
    }\\
    \vspace{-5pt}
    
    \subfloat[][\label{col_imgs:c}\centering Left to right: adversarial image formed on SpliceRadar, visualized adversarial perturbation, heatmaps via EXIF-SC, SpliceRadar, and Noiseprint]{
    %\hspace{-5pt}
    \ic[.19]{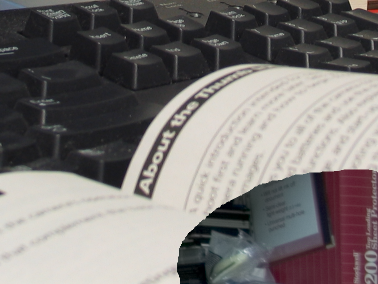} \hspace{-1pt}
    \ic[.19]{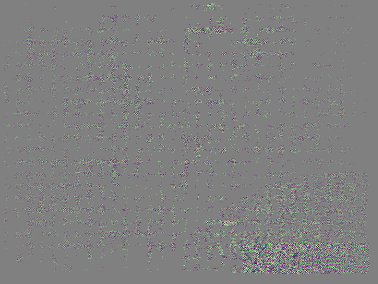} \hspace{-1pt}
    \ic[.19]{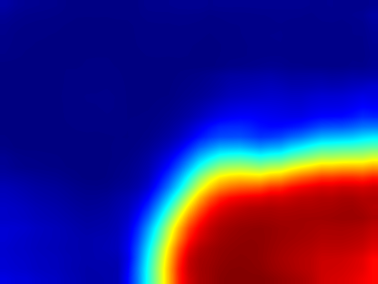} \hspace{-1pt}
    \ic[.19]{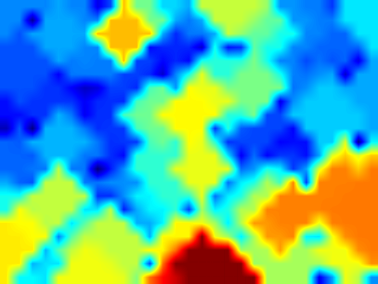} \hspace{-1pt}
    \ic[.19]{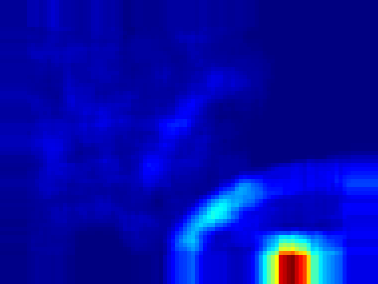}
    }\\
    \vspace{-5pt}
  
    \subfloat[][\label{col_imgs:d}\centering Left to right: adversarial image formed on Noiseprint, visualized adversarial perturbation, heatmaps via EXIF-SC, SpliceRadar, and Noiseprint]{
    %\hspace{-5pt}
    \ic[.19]{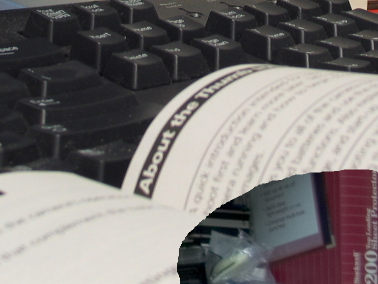} \hspace{-1pt}
    \ic[.19]{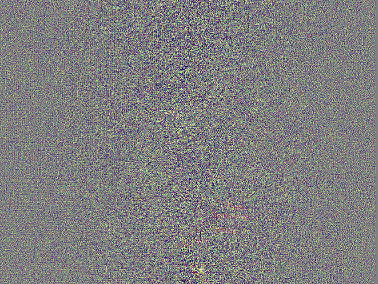} \hspace{-1pt}
    \ic[.19]{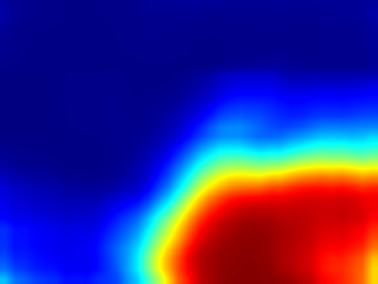} \hspace{-1pt}
    \ic[.19]{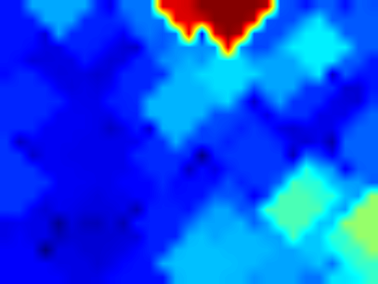} \hspace{-1pt}
    \ic[.19]{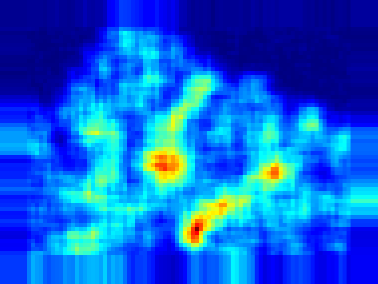}
    }\\
    \vspace{-5pt}
    
    \cap{fig:col_imgs}{Adversarial Examples on the Columbia Dataset.}{This figure presents the qualitative results of adversarial white-box and black-box attacks compared to the baseline performance of EXIF-SC, SpliceRadar, and Noiseprint on a spliced image. The visualized heatmaps are scaled; blue for lowest and red for highest values. Note that adversarial perturbations are depicted on a gray background and magnified by a factor of 100 for better visualization.}
    
    \vspace{1pt}
    %\vspace{-4pt}
\end{figure*}

\begin{figure*}
    \centering
    
    \subfloat[][\label{dso_imgs:a}\centering Left to right: spliced image, ground-truth manipulation mask, heatmaps via EXIF-SC, SpliceRadar, and Noiseprint]{
    %\hspace{-5pt}
    \ic[.19]{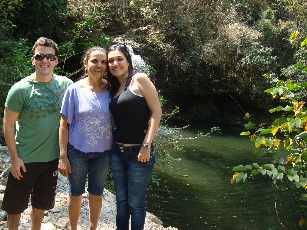} \hspace{-1pt}
    \ic[.19]{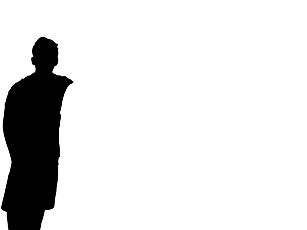} \hspace{-1pt}
    \ic[.19]{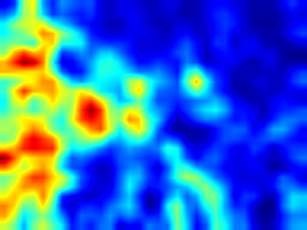} \hspace{-1pt}
    \ic[.19]{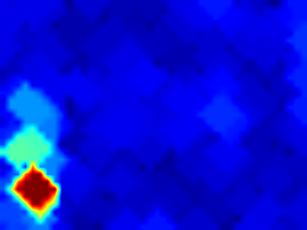} \hspace{-1pt}
    \ic[.19]{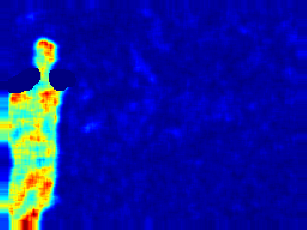}
    }\\
    \vspace{-5pt}
    
    \subfloat[][\label{dso_imgs:b}\centering Left to right: adversarial image formed on EXIF-SC, visualized adversarial perturbation, heatmaps via EXIF-SC, SpliceRadar, and Noiseprint]{
    %\hspace{-5pt}
    \ic[.19]{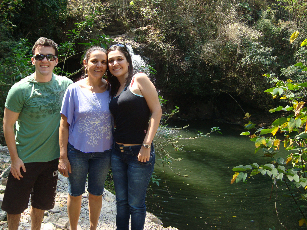} \hspace{-1pt} \ic[.19]{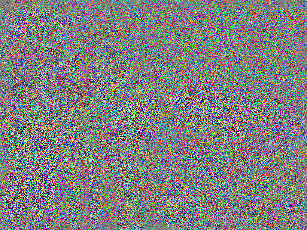} \hspace{-1pt} \ic[.19]{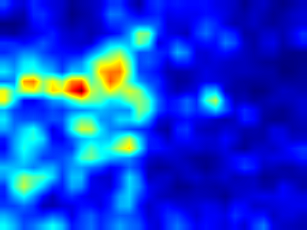} \hspace{-1pt} \ic[.19]{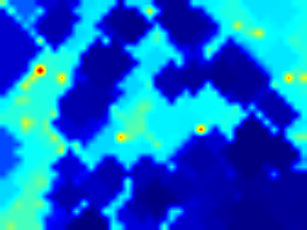} \hspace{-1pt}
    \ic[.19]{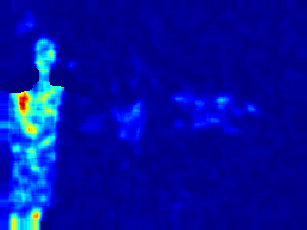}    
    }\\
    \vspace{-5pt}
    
    \subfloat[][\label{dso_imgs:c} Left to right: adversarial image formed on SpliceRadar, visualized adversarial perturbation, heatmaps via EXIF-SC, SpliceRadar, and Noiseprint]{
    %\hspace{-5pt}
    \ic[.19]{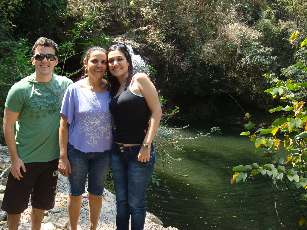} \hspace{-1pt} \ic[.19]{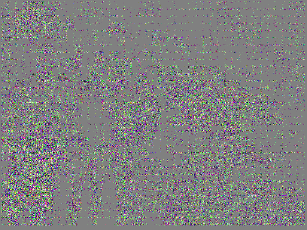} \hspace{-1pt} \ic[.19]{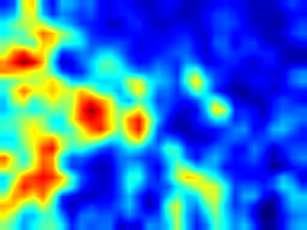} \hspace{-1pt} \ic[.19]{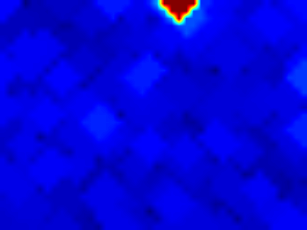} \hspace{-1pt}
    \ic[.19]{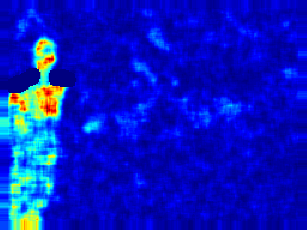}     
    }\\
    \vspace{-5pt}

    \subfloat[][\label{dso_imgs:d} Left to right: adversarial image formed on Noiseprint, visualized adversarial perturbation, heatmaps via EXIF-SC, SpliceRadar, and Noiseprint]{
    %\hspace{-5pt}
    \ic[.19]{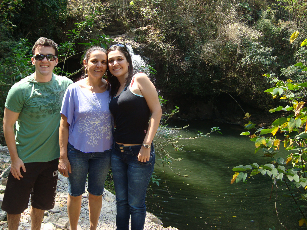} \hspace{-1pt} \ic[.19]{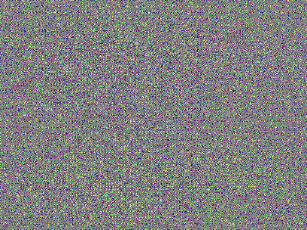} \hspace{-1pt} \ic[.19]{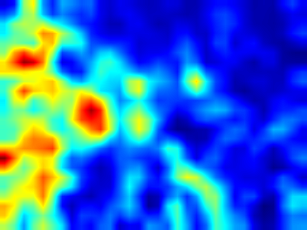} \hspace{-1pt} \ic[.19]{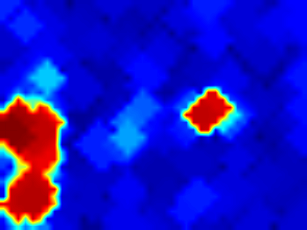} \hspace{-1pt}
    \ic[.19]{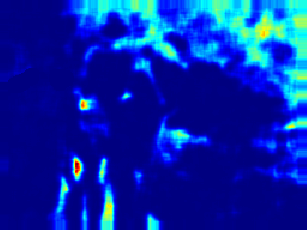}     
    }\\
    \vspace{-5pt}
    
    \cap{fig:dso_imgs}{Adversarial Examples on the DSO-1 Dataset.}{This figure demonstrates the qualitative results of adversarial white-box and black-box attacks compared to the baseline performance of EXIF-SC, SpliceRadar, and Noiseprint on a spliced image. The visualized heatmaps are scaled; blue for lowest and red for highest values. Note that adversarial perturbations are depicted on a gray background and magnified by a factor of 100 for better visualization.}

    %\vspace{0pt}
    \vspace{2pt}
\end{figure*}

\begin{figure*}
    \centering
    
    \subfloat[][\label{nc_imgs:a}\centering Left to right: spliced image, ground-truth manipulation mask, heatmaps via EXIF-SC, SpliceRadar, and Noiseprint]{
    %\hspace{-5pt}
    \ic[.19]{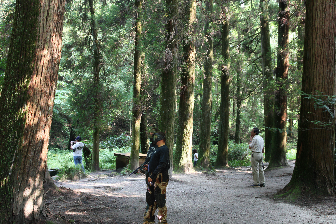} \hspace{-1pt}
    \ic[.19]{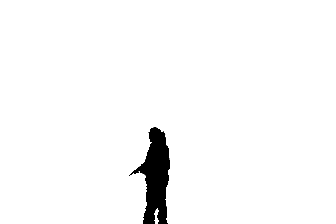} \hspace{-1pt}
    \ic[.19]{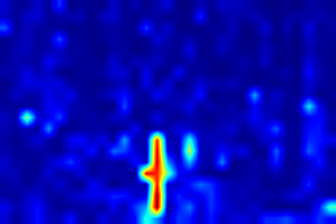} \hspace{-1pt}
    \ic[.19]{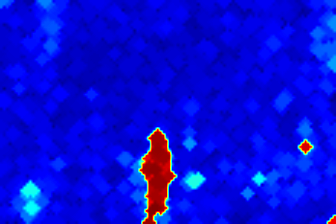} \hspace{-1pt}
    \ic[.19]{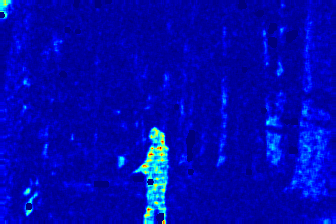}
    }\\
    \vspace{-4pt}
    
    \subfloat[][\label{nc_imgs:b}\centering Left to right: adversarial image formed on EXIF-SC, visualized adversarial perturbation, heatmaps via EXIF-SC, SpliceRadar, and Noiseprint]{
    %\hspace{-5pt}
    \ic[.19]{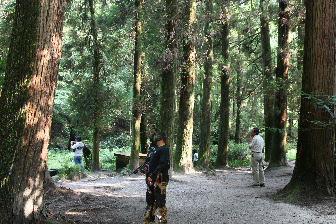} \hspace{-1pt} \ic[.19]{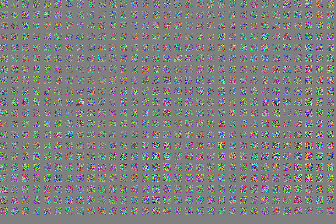} \hspace{-1pt} \ic[.19]{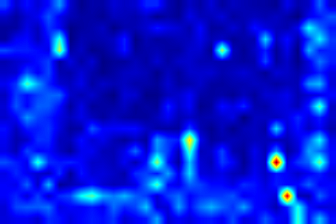} \hspace{-1pt}
    \ic[.19]{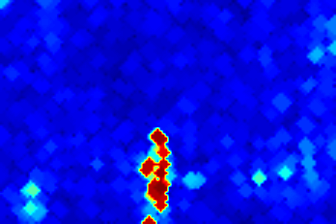} \hspace{-1pt}
    \ic[.19]{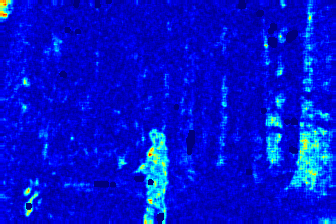}    
    }\\
    \vspace{-4pt}
    
    \subfloat[][\label{nc_imgs:c} Left to right: adversarial image formed on SpliceRadar, visualized adversarial perturbation, heatmaps via EXIF-SC, SpliceRadar, and Noiseprint]{
    %\hspace{-5pt}
    \ic[.19]{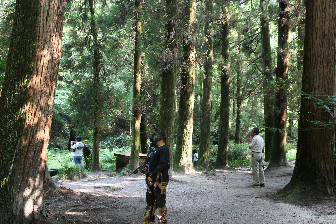} \hspace{-1pt} \ic[.19]{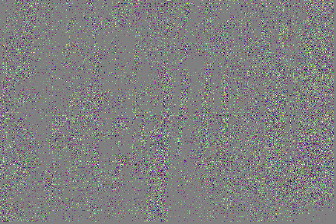} \hspace{-1pt} \ic[.19]{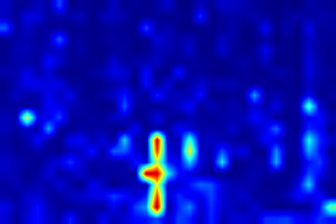} \hspace{-1pt}
    \ic[.19]{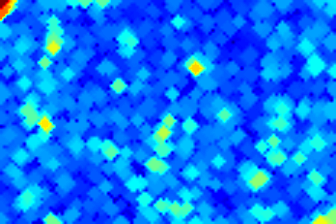} \hspace{-1pt}
    \ic[.19]{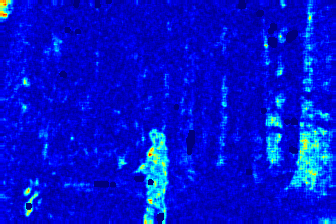}     
    }\\
    \vspace{-4pt}

    \subfloat[][\label{nc_imgs:d} Left to right: adversarial image formed on Noiseprint, visualized adversarial perturbation, heatmaps via EXIF-SC, SpliceRadar, and Noiseprint]{
    %\hspace{-5pt}
    \ic[.19]{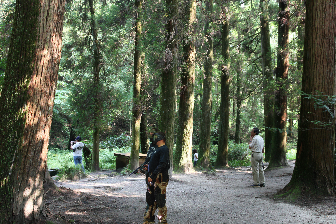} \hspace{-1pt} \ic[.19]{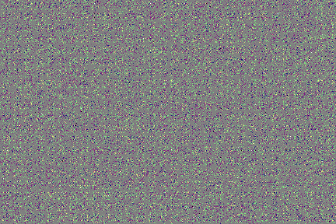} \hspace{-1pt} \ic[.19]{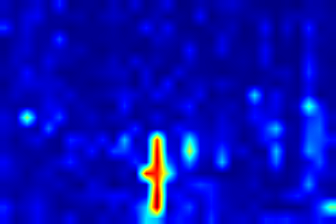} \hspace{-1pt} \ic[.19]{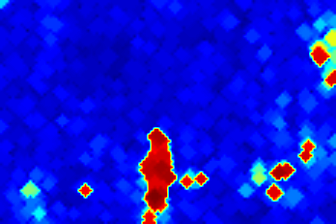} \hspace{-1pt}
    \ic[.19]{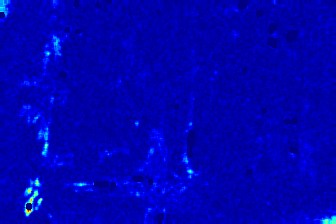}     
    }\\
    %\vspace{-4pt}

    \cap{fig:nc_imgs}{Adversarial Examples on the NC16 Dataset.}{This figure demonstrates the qualitative results of adversarial white-box and black-box attacks compared to the baseline performance of EXIF-SC, SpliceRadar, and Noiseprint on a spliced image. The visualized heatmaps are scaled; blue for lowest and red for highest values. Note that adversarial perturbations are depicted on a gray background and magnified by a factor of 100 for better visualization.}

    \vspace{1pt}
    %\vspace{-5pt}
\end{figure*}
%\vfill

\textbf{Evaluation metrics.}
To evaluate the adversarial robustness of the three targeted image-forensics methods, we compute and compare four performance measures that are commonly used in the literature \cite{huh2018fighting,ghosh2019spliceradar,cozzolino2019noiseprint} -- F1 score, the Matthews Correlation Coefficient (MCC), the Mean Average Precision (MAP), and the area under the Receiving Operating Characteristic curve (ROC-AUC or AUC, in short) -- before and after the adversarial attack takes place.
Note that both F1 and MCC operate on binary decision maps, while forgery localization tools typically provide a probability map as output, which is then converted into a binary map using a threshold.
Regarding the selection of the optimal threshold value, which is utilized to distinguish genuine and forged regions based upon the formed heatmap, we follow the same procedure as described in \cite{huh2018fighting, ghosh2019spliceradar}.

\subsection{Results}

The quantitative results of our experiments are summarized in Table \ref{tab:results}.
To demonstrate the qualitative properties of our approach, we show examples for the Columbia, DSO-1, and NC16 datasets in Figures \ref{fig:col_imgs}, \ref{fig:dso_imgs}, and \ref{fig:nc_imgs}, respectively.
As we will see, on the more challenging NC16 dataset the tested image-forensics methods do poorly.
Since attacking an already struggling splice localization tool does not have much space for making further damage, we also present results on a subset containing the manipulated images for which EXIF-SC, SpliceRadar, and Noiseprint perform better regarding the localization of the manipulated areas (F1$\geq$0.6).
The results for this subset consisting of 83 images of the NC16 dataset are presented in Table \ref{tab:nc16_sub}.

\textbf{Baselines.}
As a starting point, we obtain the performance measures for the three targeted image-forensics algorithms on the forged images of the three datasets.
We can see in Table \ref{tab:results} that EXIF-SC outperforms SpliceRadar and Noiseprint on the Columbia dataset but falls behind them on the more challenging DSO-1 and NC16 datasets.
In Figures \ref{fig:col_imgs}\subref{col_imgs:a}, \ref{fig:dso_imgs}\subref{dso_imgs:a}, \ref{fig:nc_imgs}\subref{nc_imgs:a}, we show the qualitative splice localization performances of EXIF-SC, SpliceRadar, and Noiseprint on spliced images of the Columbia, DSO-1, and NC16 datasets, respectively.
As we can see, the three image-forensics methods can more or less successfully localize the manipulated regions denoted by the manipulation masks as the authentic and forged regions can be separated on the computed heatmaps by applicable threshold values.
Note that EXIF-SC also produces some false positive splice localization on the DSO-1 example shown in Figure \ref{fig:dso_imgs}\subref{dso_imgs:a}.

\textbf{White-box attacks.}
To put the results of our adversarial white-box attacks on EXIF-SC, SpliceRadar, and Noiseprint into perspective, we show the applicable performance measures for an ``ideal attack'' in the last row of Table \ref{tab:results} labeled as \textit{All Authentic Attack}.
In short, this ideal attack occurs when each pixel -- both authentic and manipulated -- of the manipulated image is declared as authentic by the image-forensics tool.
Note that this ideal attack does not yield lower bounds regarding the performance measures that we can achieve.
Due to the fact that SpliceRadar and Noiseprint utilize a Gaussian Mixture Model (GMM) with two components, they always declare authentic and manipulated regions.
Consequently, it is impossible to perform an ideal attack on these two image-forensics tools.
Compared to baselines, we can see in Table \ref{tab:results} that white-box attacks greatly decrease the performance measures of both EXIF-SC and SpliceRadar, and we can notice that Noiseprint is more resilient, especially on the Columbia dataset.
The qualitative results depicted in Figures \ref{fig:col_imgs}\subref{col_imgs:b}-\subref{col_imgs:d}, \ref{fig:dso_imgs}\subref{dso_imgs:b}-\subref{dso_imgs:d}, and \ref{fig:nc_imgs}\subref{nc_imgs:b}-\subref{nc_imgs:d} (leftmost images) demonstrate that the formed adversarial examples contain imperceptibly small perturbations that are capable of impeding the targeted image-forensics methods.
Considering the visualized perturbations, we notice that those formed on EXIF-SC are stronger than the others generated on SpliceRadar and Noiseprint.
Note that due to our algorithm mimicking the target representation with each patch, the false positive splice localizations via EXIF-SC for the DSO-1 example are all declared as authentic regions for the adversarially perturbed image shown in Figure \ref{fig:dso_imgs}\subref{dso_imgs:b} (third image).
Overall, the qualitative results visually demonstrate that the adversarial white-box attacks have succeeded as authentic and manipulated regions cannot be separated anymore via the targeted splice localization tools. 

\textbf{Transferability of adversarial examples.}
Using the adversarial examples that we have generated via white-box attacks, we can also perform adversarial black-box attacks by simply evaluating those examples on the other image-forensics tools.
This way we can assess the transferability of the formed adversarial perturbations and study the feasibility of such black-box attacks against EXIF-SC, SpliceRadar, and Noiseprint.
Note that the vulnerability to black-box attacks is more concerning from a security perspective as adversaries can impede their targeted system without having direct access to it.
The results in Table \ref{tab:results} highlight an asymmetric behavior; we can see that while adversarial perturbations formed on EXIF-SC cause a significant deterioration in the localization performance of SpliceRadar, perturbations generated on SpliceRadar barely make a difference on EXIF-SC compared to its baseline performance.
While EXIF-SC appears to be less susceptible to examples formed by adversarial white-box attacks on other image-forensics methods, the forgery localization performance of Noiseprint noticeably deteriorates under such black-box attacks.
Interestingly, considering the results on the Columbia dataset we find that black-box attacks are slightly more successful on Noiseprint than the white-box attacks.
Regarding qualitative results, we can see in Figures \ref{fig:col_imgs}\subref{col_imgs:b} and \ref{fig:dso_imgs}\subref{dso_imgs:b} that the adversarial perturbations formed on EXIF-SC limit the splice localization capabilities of SpliceRadar.
While the other adversarially perturbed Columbia, DSO-1, and NC16 images shown in Figures \ref{fig:col_imgs}-\ref{fig:nc_imgs} are not fully transferable to other image-forensics tools, some yield predictions of smaller manipulated regions, e.g., Noiseprint for the adversarial examples formed on EXIF-SC and SpliceRadar presented in Figure \ref{fig:col_imgs}\subref{col_imgs:b}-\subref{col_imgs:c} (rightmost images).

\begin{table*}[ht!]
\footnotesize
\begin{center}
\begin{tabular}{|l|ccc|ccc|ccc|}
\hline
LOCALIZATION METHOD & \multicolumn{3}{c|}{COLUMBIA} & \multicolumn{3}{c|}{DSO-1} & \multicolumn{3}{c|}{NC16} \\
(results on perturbed images) & $l_0$ & $l_2$ & $l_\infty$ & $l_0$ & $l_2$ & $l_\infty$ & $l_0$ & $l_2$ & $l_\infty$\\
\hline\hline
EXIF-SC & 1109644.39 & 1877.47 & 19.86 & 5263130.16 & 3318.37 & 18.16 & 9012808.50 & 3592.89 & 15.05 \\
SpliceRadar & \phantom{0}568532.84 & \phantom{0}981.04 & 12.06 & 1828703.85 & 1625.76 & \phantom{0}9.98 & 3195024.54 & 1606.36 & \phantom{0}9.07 \\
Noiseprint & \phantom{0}681162.14 & 1338.88 & 22.40 & 1762103.27 & 1554.88 & 24.88 & 3464998.91 & 1599.90 & 15.50 \\
\hline
\end{tabular}
%\vspace{-5pt}
\end{center}
\cap{tab:noises}{Perturbation Metrics.}{This table shows the means for various norms of the adversarial perturbations that we obtained while attacking EXIF-SC, SpliceRadar, and Noiseprint using the manipulated images of the Columbia, DSO-1, and NC16 datasets.}
%\cap{tab:noises}{Metrics of Adversarial Perturbations.}{This table shows the means for various norms of adversarial perturbations that we generated via the adapted LOTS while attacking EXIF-SC, SpliceRadar, and Noiseprint on the manipulated images of the Columbia, DSO-1, and NC16 datasets.}

\vspace{-11pt}
\end{table*}

\section{Discussion}

As we have seen, adversarial white-box attacks can generally be considered successful on both EXIF-SC and SpliceRadar.
However, the results of the white-box attacks on Noiseprint and the asymmetric transferability of adversarial examples require further analysis.

\subsection{On the robustness of Noiseprint}

Compared to EXIF-SC and SpliceRadar, the results indicate that Noiseprint is more resilient to our adversarial white-box attack, especially on the Columbia dataset.
To make sure that the higher resistance is not due to our choice of mimicking a target defined as the mean representation of authentic patches, we have conducted image-based adversarial white-box attacks on the Columbia dataset that also contains authentic counterparts of the spliced images.
Namely, instead of manipulating patches in order to move their representations closer to the predefined target, for each spliced image we have identified their authentic counterpart, extracted the feature representation from that image, and used it as our target to mimic with the feature representation extracted from the whole corresponding spliced image.
Overall, this image-based white-box attack has failed to outperform our patch-based approach.

We conjecture that the higher robustness of Noiseprint compared to EXIF-SC and SpliceRadar is due to the higher dimensional features extracted from manipulated images and the lack of bottlenecks in the network architecture used for feature extraction.
Both bottlenecks and lower dimensional features yield possible collisions -- patches or images having the same extracted feature representations -- which can be exploited by adversaries.
While both EXIF-SC and SpliceRadar work with feature representations of lower dimensionalities -- with default settings, $72\times72$-pixel patches projected into 100-dimensional and $128\times128$-pixel patches mapped into 4096-dimensional feature representations, respectively -- Noiseprint yields 2-D noise residual maps with height and width identical to its inputs.

\subsection{Asymmetric transferability}

To understand the asymmetric transferability properties of adversarial examples, especially between EXIF-SC and SpliceRadar, we have analyzed the perturbations that we formed on the three targeted image-forensics algorithms.
The quantitative results are summarized in Table \ref{tab:noises}.
We can see that the adversarial perturbations formed by our approach while attacking EXIF-SC are stronger than those obtained on SpliceRadar or Noiseprint.
Adversarial attacks on EXIF-SC affect the most pixels, as indicated by $l_0$ norms, and they lead to the highest Euclidean norms as well.

Assuming that stronger perturbations yield higher transferability rates is a reasonable hypothesis.
To validate that, we have formed new examples by scaling the formed perturbations; we have doubled the perturbations of examples that we obtained on SpliceRadar and halved those we generated while attacking EXIF-SC.
After evaluating these new examples, we have found that the transferability of adversarial examples is not strictly determined by the strength of perturbations in terms of $l_0$, $l_2$, or $l_\infty$ norms.
Considering the results on the subset of NC16 in Table \ref{tab:nc16_sub}, we can see on both SpliceRadar and Noiseprint that stronger perturbations formed on EXIF-SC cause lower deterioration on the splice localization performance than weaker perturbations generated on the other two image-forensics methods.

\vfill
\section{Conclusion}

Image forensics can play an important role to prevent the spread of manipulated content on communication channels, such as digital and social media.
The detection and/or localization of manipulated regions on forged images have already attracted the research community, and this research area will likely gain more focus in the future thanks to the increasing demand for such tools.
Due to recent advances and trends in machine learning, researchers have proposed deep learning-based image-forensics methods that perform well regarding the localization of various manipulations, including splices.
However, the applied deep learning models are inherently susceptible to adversarial perturbations making such tools vulnerable as well.

In this paper, we have demonstrated on three image-forensics methods, EXIF-SC, SpliceRadar, and Noiseprint, that splice localization via non end-to-end deep learning-based approaches can be impeded by directly manipulating the internal representations of image patches.
Furthermore, we have found that adversarial perturbations can be transferable among the tested image-forensics methods to decrease their capabilities with different effectiveness.

In summary, deep learning-based image-forensics methods that were designed to reveal manipulations, ironically, can be impeded by the application of additional manipulations.
To avoid this problem, we need deep learning models that are inherently robust to adversarial perturbations.

%\vfill
\pagebreak

{\small
\bibliographystyle{ieee_fullname}
\bibliography{egbib}
}

\end{document}